\documentclass{article}

\usepackage{arxiv}

\usepackage[utf8]{inputenc} 
\usepackage[T1]{fontenc}    
\usepackage{hyperref}       
\usepackage{url}            
\usepackage{booktabs}       
\usepackage{amsfonts}       
\usepackage{nicefrac}       
\usepackage{microtype}      
\usepackage{lipsum}
\usepackage{graphicx}
\graphicspath{ {./images/} }

\title{Public Transport Network Design for Equality of Accessibility via Message Passing Neural Networks and Reinforcement Learning}

\author{
 Duo Wang \\
  SAMOVAR, Télécom SudParis, Institut Polytechnique de Paris\\
  \texttt{duo-wang@telecom-sudparis.eu} \\
   \And
 Maximilien Chau \\
  SAMOVAR, Télécom SudParis, Institut Polytechnique de Paris\\
  \texttt{maximilien.chau@ip-paris.fr} \\
  \And
 Andrea Araldo \\
  SAMOVAR, Télécom SudParis, Institut Polytechnique de Paris\\
  \texttt{andrea.araldo@telecom-sudparis.eu} }

\usepackage{algorithm}
\usepackage{algorithmic}
\usepackage{amsmath}
\usepackage{graphicx}
\usepackage{subfig}
\usepackage{subcaption}
\DeclareMathOperator*{\argmax}{argmax}

\AtBeginDocument{%
  \providecommand\BibTeX{{%
    \normalfont B\kern-0.5em{\scshape i\kern-0.25em b}\kern-0.8em\TeX}}}

\newcommand{\keepcomment}{0} 
\usepackage[normalem]{ulem}
\ifnum\keepcomment=1
	\usepackage[colorinlistoftodos,textsize=scriptsize]{todonotes} 
    \newcommand{\stkout}[1]{\ifmmode\text{\sout{\ensuremath{#1}}}\else\sout{#1}\fi}
    
    \newcommand\todoi[1]{\textcolor{orange}{#1}}
\else
	
	\usepackage[disable]{todonotes} 
    \newcommand\todoi[1]{}
\fi

\usepackage{calrsfs} 
\DeclareMathAlphabet{\pazocal}{OMS}{zplm}{m}{n}

\begin{document}
\maketitle

\bibliographystyle{unsrt}

\begin{abstract}
Designing Public Transport (PT) networks able to satisfy mobility needs of people is essential to reduce the number of individual vehicles on the road, and thus pollution and congestion. Urban sustainability is thus tightly coupled to an efficient PT. Current approaches on Transport Network Design (TND) generally aim to optimize generalized cost, i.e., a unique number including operator and users' costs. Since we intend quality of PT as the capability of satisfying mobility needs, we focus instead on PT accessibility, i.e., the ease of reaching surrounding points of interest via PT. PT accessibility is generally unequally distributed in urban regions: suburbs generally suffer from poor PT accessibility, which condemns residents therein to be dependent on their private cars.  We thus tackle the problem of designing bus lines so as to minimize the inequality in the geographical distribution of accessibility. We combine state-of-the-art Message Passing Neural Networks (MPNN) and Reinforcement Learning. We show the efficacy of our method against metaheuristics (classically used in TND) in a use case representing in simplified terms the city of Montreal. 
\end{abstract}


\section{Introduction}
\label{Introduction_}

Existing Public Transport (PT) is less and less adequate to satisfy mobility needs of the people, in a context of urban sprawl~\cite{sun2018optimal}. The United Nations estimate that only ``1/2 of the urban population has convenient access to PT~''\cite{Un_PT}. 
Building more and more PT lines to keep pace with urban sprawl, using traditional planning objectives, has proved to be ineffective.

PT operators generally design PT lines with the purpose of maximizing overall efficiency, measured in terms of generalized cost (which takes into account travel times and cost for the operators), or number of kilometers traveled or number of passengers transported. This has resulted in unequal development of PT within urban areas. The level of service offered by PT is often satisfying in city centers and poor in suburbs. Therefore, suburban population  depend on their private cars to perform their daily activities~\cite{anable2005complacent,WELCH201329}. As an example, the modal share of car in the city center of Prague is double than in the city center.
\todo{\\
aa: A citation is needed to support this data\\}
The dependence on private cars has negative economic, social, and environmental impacts (\cite[Section~2.2]{Boussauw2022}), which are common to different cities of the world. For example, 61\% of EU road transport CO2 comes from cars, jobseekers with no car have 72\% less chances of finding a job in Flanders, etc. Therefore, a sufficient condition to achieve sustainability is to improve PT level of service where it is currently poor. We propose in this paper to set geographical equality of PT level of service as the main design objective. We focus in this paper on PT \emph{accessibility} metric, which measures the ease (in terms of time and/or monetary cost) of reaching Points of Interest (PoIs) via PT. To improve geographical equality, we prioritize increasing PT accessibility in the underserved areas.

A trivial strategy to do so would be to place more stops and lines in underserved areas. However, this may not be the most efficient way to increase accessibility there. Indeed, the ability to reach PoIs might be increased even more by improving PT network close to other nodes, possibly far away, that may enable convenient changes with other important lines. In general the PT network extends the inter-dependencies between locations and PoIs far beyond those that are in physical proximity. Therefore, to reduce the inequality of the distribution of accessibility it is always required to take the entire PT graph into consideration, rather than just around the local areas where we want improvement. This makes our problem particularly challenging.

Most cities already have an existing PT network and the need to build lines from scratch is very limited. Re-designing the whole PT network is also not an option as it would lead to major costs to the operators. For this reason, in this work, we assume a core PT network (e.g., metro) that does not change and we only tackle the design of some bus lines, which comes at a limited infrastructure cost. By focusing on bus network design only, we aim to achieve important reduction of accessibility inequality with relatively low expenses for the operator.

The contribution of this paper is a novel approach to PT network design, in which the non-trivial inter-dependencies involved into the accessibility metrics are captured via a Message Passing Neural Network (MPNN)~\cite{LeiChen2023,gilmer2017neural,Kutyniok2022} and a Deep Reinforcement Learning (RL) agent. While MPNN and RL have been used to solve canonical optimization problems on graphs, to the best of our knowledge we are the first to use them for PT network design. To reduce inequality, we propose a simple yet effective approach, consisting in using quantiles of the accessibility metrics as objective function.

Numerical results in a scenario inspired by Montreal show that our method effectively reduces accessibility inequality, more effectively than metaheuristics classically used for PT design. This improvement is due to the capability of the MPNN to capture the structure of the PT network and its relation with the PoIs, while metaheuristics do no learn any dependencies and restrict themselves in randomly exploring the space of designs.

\section{Related work}
\label{Related_work_}


Transport Network Design Problems (TNDPs), and in particular Public Transport Network Design Problems (PTNDPs) can be at a strategic level or an operational level. At a strategic level, a PT planner aims to decide the route of the different lines as well as their frequencies. At an operational level, a PT operator organizes the service in order to match the decisions taken at the strategic level, deciding precise time tables, as well as crew and vehicle scheduling. We focus in this paper on PTNDPs at a strategic level. A review of strategic level PTNDPs is provided in~\cite{farahani2013review}. The methods generally used to solve PTNDPs can be divided into two categories:  mathematical programming methods~(Section~\ref{sec:mathematical-programming}) and search-based heuristics methods~(Section~\ref{sec:heuristic-methods}). We use instead graph-based reinforcement learning~(Section~\ref{sec:graph-based-reinforcement-learning}). The latter has been applied to solve several combinatorial problems and has also few applications in Transport. However, it has not been used for PT planning (lines design), with one exception~\cite{yoo2023reinforcement},with respect to whom we will pinpoint the difference with respect to our work.

\subsection{Mathematical programming methods}
\label{sec:mathematical-programming}
TNDPs usually be formulated as non-linear programming models. Solvers are used to solve these models. To ensure that their models are reasonable, they usually need to set numerous constraints \cite{wei2021optimal}. For a realistic sized problem, it is difficult to find a suitable solution with this method \cite{chakroborty2003genetic}. Therefore, instead of considering a real city, they turned to representing it with a regular geometrical pattern, such as Continuous Approximation method \cite{calabro2023adaptive}. This method can indeed crudely describe any city through some characteristics, but in the end they cannot be applied to any real city. Because real cities are much more complex than abstract regular geometrical shapes. 

Some works try to achieve the purpose of dealing with large-scale real cities by adding more constraints, which often limit the solution space. These works are \cite{gutierrez2018corridor}, which first proposed to design metro lines in some predefined corridors, and these corridors with higher passenger traffic were chosen by a Greedy generation heuristic step before any optimization and \cite{wei2019strategic}, conducted a real-world case study by predefining corridors and solving a bi-objective mixed-integer linear programming model. However, all these works highly rely on an expert guidance to help reduce the potential solution space, in other words, the results vary with different expert guidance. Expert guidance does not ensure that the optimal solution is not eliminated.

\subsection{Heuristics methods}
\label{sec:heuristic-methods}

The most commonly used search-based heuristic methods are Simulated Annealing, Tabu Search and Genetic Algorithm. Their common method is to initialize a PT design, and then gradually optimize the PT design by changing it through heuristics. \cite{schittekat2013metaheuristic} develops a bilevel mataheuristic to find the optimal solution of school bus routing problem. The upper level repeats greedy randomized adaptive search followed by a variable neighborhood descent $n_{max}$ times and then finds the best solution to bus stops assignment among these $n_{max}$ times of search, while the lower level finds an exact solution to a sub-problem of assigning students to stops by solving a mathematical programming problem. \cite{owais2018complete} uses a Genetic Algorithm
to generate a bus route network from With only the Origin–Destination matrix and the network structure of an existing transportation network. In the process of using GA to evolve route design, the connectivity of routes is ensured at every stage of the GA. These methods have a common limitation, that is, during the iterative process, some basic characteristics of the graph are retained, such as connectivity, which is not needed in reality. Therefore, they often can only find local optimal solutions.

Considering all of the above, we need to design a generic algorithm which requires fewer constraints. It does not require expert experience, which is the limitation of Mathematical programming method, and it also can no longer retain some unreasonable characteristics of graph, which is the limitation of Search-based heuristics method. Therefore, we chose the RL-based method. 


Closer to our work, the work in \cite{yoo2023reinforcement} solved TNDP by directly using RL, and did not extract the information of PT graph. The work \cite{darwish2020optimising} used Transformer architecture to produce the nodes and the graph embeddings, and then solved TNDP via RL. But to make their method feasible, they assumed that the network is a connected graph, which is not needed in our paper. The work \cite{wei2020city} presented a RL-based method to solve the city metro network expansion problem. Our main difference lies in the different methods used to extract the information on PT graph. They used two 1-dimensional convolutional neural networks to calculate the embeddings for the stations. However, the graph structure is complex, we believe only applying 1-dimensional convolutional neural network is insufficient. In this sense, we proposed to use Message Passing Neural Network (MPNN) to extract the information on PT graph.

\subsection{Graph-based Reinforcement Learning applications}
\label{sec:graph-based-reinforcement-learning}

Some works using Graph-based RL to solve other problems are as follows:
The work \cite{barrett2020exploratory} already coupled it successfully to a DQN on traditional combinatorial problem, such as max-cut problem. The work \cite{duan2020efficiently} first extracted the information of graphs, based on the recurrent neural networks (RNN), and then combined it with RL to solve the Vehicle Routing Problem. The work \cite{yoon2021transferable} improved the transferability of the solution to traffic signal control problem by combining Reinforcement learning and MPNN. The work \cite{koksal2022reinforcement} proposed an algorithm for the vehicle fleet scheduling problem, by integrating a reinforcement learning approach with a genetic algorithm. The reinforcement learning is used to decide parameters of genetic algorithm.

\section{Model}
\label{Model_}


\begin{table}[]
\centering
  \caption{Table of Notation}
  \label{tab:ppp}
  \begin{tabular}{cc}
    \toprule
    PT &Public Transport (\S\ref{Introduction_})\\
    RL-based Equality algorithm &RLEA (\S\ref{Introduction_})\\
    Message Passing Neural Network & MPNN (\S\ref{Introduction_})\\
    Deep Q-learning & DQN (\S\ref{Introduction_})\\
    Transportation Network Design Problems &TNDPs (\S\ref{Related_work_})\\

    PoIs & Points of Interest (\S\ref{territoryPT_})\\
    Markov Decision Problem & MDP (\S\ref{sec:RL_formulation})\\
    \hline
    $\pazocal{G}$ & Graph of PT network (\S\ref{territoryPT_})\\
    $\pazocal{P}$ & Set of all PoIs (\S\ref{Accessibility_})\\
    $\pazocal{B}$ & Set of candidate nodes (\S\ref{sec:optimization-problem})\\ 
    $\overline{\pazocal{B}}$ & Set of non candidate nodes (\S\ref{sec:optimization-problem})\\ 
    $\pazocal{C}$ & Set of all centroids (\S\ref{Accessibility_})\\
    $\pazocal{C}^{q}$ & Set the q\% of centroids \\
                      & with the worst accessibility (\S\ref{Accessibility_})\\
    \hline
    $T_\text{max}$ & Maximum accessibility time (\S\ref{Accessibility_})\\
    $n_b$ & Number of all candidate \\
                      & bus stops (\S\ref{sec:optimization-problem})\\
    $k$ & Number of new lines (\S\ref{sec:optimization-problem})\\
    $l_i$ & Bus line $i$ (\S\ref{sec:optimization-problem})\\
    $b_i$ & Bus stop $i$ (\S\ref{sec:RL_formulation})\\
    $S$ & Partition of candidate stops (\S\ref{sec:RL_formulation})\\

    $\mu_b$ & Embedding vector of stop $b$ (\S\ref{MPNeural_Network})\\
    $x_b$ & Feature vector stop $b$ (\S\ref{MPNeural_Network})\\
    $X$ & Matrix of feature vectors (\S\ref{MPNeural_Network})\\
    $M_\text{adj}$ & Bus stop adjacency matrix (\S\ref{MPNeural_Network})\\
    $\pazocal{N}(b)$ & Neighbors of bus stop $b$ (\S\ref{MPNeural_Network})\\
    $Q(\cdot)$  & State-value function (\S\ref{MPNeural_Network})\\
    $M(\cdot)$  & Message function (\S\ref{MPNeural_Network})\\
    $U(\cdot)$ & Update function (\S\ref{MPNeural_Network})\\
    $R(\cdot)$ & Readout function (\S\ref{MPNeural_Network})\\
    $\Theta$ & Set of learning parameters (\S\ref{MPNeural_Network})\\
    $\gamma$ & Discount factor (\S\ref{MPNeural_Network})\\
    \bottomrule
  \end{tabular}
\end{table}

\begin{figure}[h]
  \centering
  \includegraphics[width=0.6\linewidth]{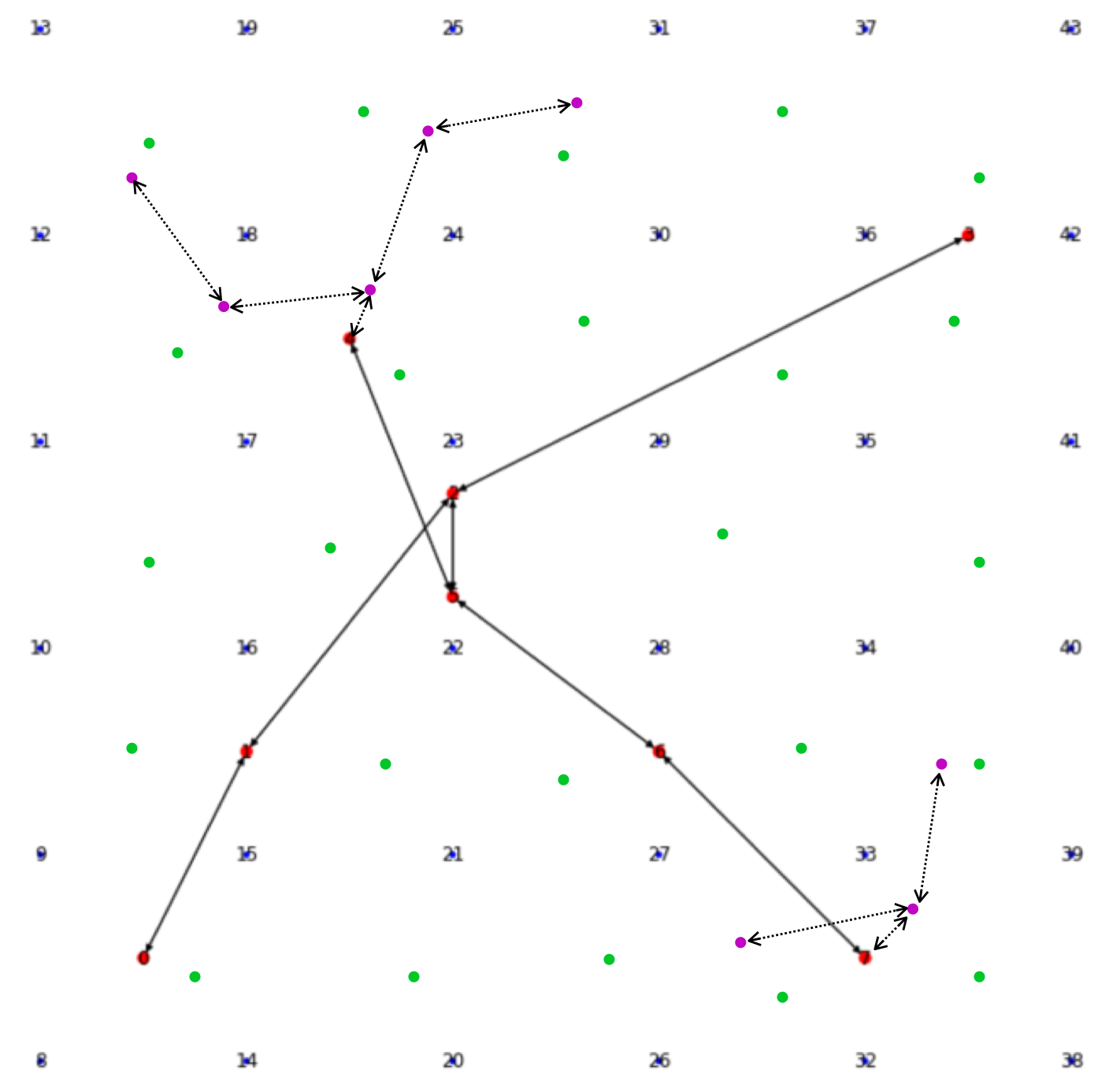}
  \caption{Model of Public Transit\textit{: PT graph $\pazocal{G}$ has 2 metro lines (red points represent metro stations) and 2 bus lines (purple points represent bus stops), in addition, the blue points are the \emph{centroids} and the green points are the points of interest}} 
  \label{fig:PT}
\end{figure}
\todoi{In Figure~\ref{fig:PT} it would be good to habe the tessellation visible}

\subsection{Model of the territory and of public transport}
\label{territoryPT_}

We partition the study area with a regular tessellation (here we adopt square tiles, but any regular shape can be used), as in Figure~\ref{fig:PT}.
The center of each tile is called \emph{centroid} (blue points in Figure~\ref{fig:PT}). The study area also contains Points of Interest (PoIs) often called ''opportunities'' in the literature about accessibility. PoIs can be shops, jobs, schools, resturants, .... PoIs are depicted as green points in Figure~\ref{fig:PT}.

%
As illustrated in Figure~\ref{fig:PT}, our model of PT is composed of:
\begin{enumerate}
    \item Metro lines and metro stations (red points).
    \item Bus lines and bus stops (purple points).
\end{enumerate}

Changing metro lines is very costly and time consuming. On the other hand, redesigning bus lines requires much less infrastructure cost and can be done in shorter time. In this paper, we only focus on redesigning bus lines and keep metro lines unchanged.

We model PT as a graph $\pazocal{G}$$=(\pazocal{V}, \pazocal{E}, \pazocal{L})$, where~$\pazocal{V}$ is the set of nodes, including centroids and PT stops (both metro and bus stops), $\pazocal{E}$ is the set of edges and~$\pazocal{L}$ is the set of lines. Any PT line $l$
%
(metro line or bus line) is a sequence of PT stops, linked by edges~$e\in\pazocal{E}$. Each edge has a weight,
%
which represents the time used by a vehicle to go from a PT stop to another.  
A PT line $l$ also has a headway $t_l$, which is the time between two vehicle departures in the same direction. 
Since we only optimize bus lines, headway~$t_l$ of any metro line remains unchanged.
For metro lines, headway~$t_l$ can be obtained from real data. Instead, since we build bus lines, we need to calculate ourselves headway~$t_l$ for any bus line~$l$. Once we decide the sequence of bus stops composing~$l$,
%
we can get the total length of the line~$d_l$, to go from the first to the last stop. We assume number~$N_l$ of buses deployed on line~$l$ is fixed in advance. Denoting with~$s_b$ the bus speed, headway~$t_l$ is:
\begin{equation}
    t_l=\frac{d_l}{s_b \cdot N_l}
    \label{eq:tl}
\end{equation}
In reality, the headway could also been a bit higher, due to the time spent by the bus at the terminal before starting the next run.

We include in $\pazocal{G}$ the set of centroids $\pazocal{C}$ and the set of points of interest $\pazocal{P}$. We also include edges (in the two directions) between any centroid and all PT stops, between any point of interest and all PT stops,
and between any centroid and any point of interest.

For a trip from centroid $c$ to point of interest $poi$, a traveler can choose between different modes of travel. For example, a traveler could simply
walk to $poi$ at speed $s_w$ or walk from centroid $c$ to a PT stop (metro station or bus stop), go via PT to another stop, and from there walk to the destination $poi$. A traveler could also change from PT line $l$ to line $l'$ at a transfer that belongs to both lines. We consider an average waiting time $t_{l'}/2$ at this station.
We assume that travelers always take the shortest path, i.e., the one that allows to arrive at destination with the least time.


\subsection{Accessibility}
\label{Accessibility_}

\begin{figure}
    \centering
    \includegraphics[width=\columnwidth]{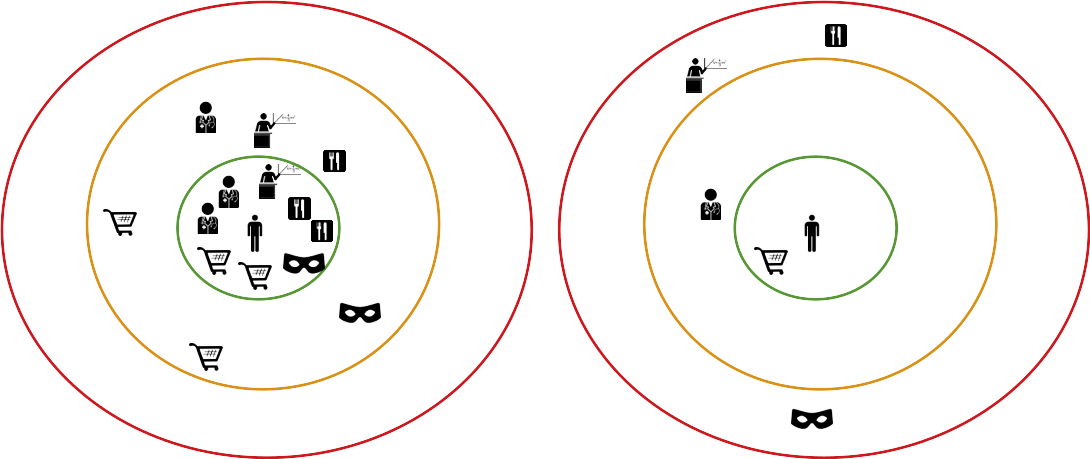}
    \caption{Accessibility example: the location on the left enjoys high accessibility as, departing from it, one can reach many PoIs in little time. On the right, instead, accessibility is poor: few PoIs are reachable and high travel times are required. The left and right locations are typical of city centers and suburbs, respectively.}
    \label{fig:accessibility}
\end{figure}

Accessibility measures the ease of reaching PoIs via PT. A simplified depiction is given in Figure~\ref{fig:accessibility}. Accessibility depends on both land use (which determines where PoIs are) and the transport system (which determines the time to reach each PoI). There are several ways of mathematically define accessibility. We define the accessibility of centroid $c$ as: 
\begin{equation}
    acc(c) = \sum_{poi \in \pazocal{P}}\max \left( 0, 1-\frac{ T_{c,poi}  }{ T_\text{max} } \right),
\label{eq:acc_centroid}
\end{equation}

where $T_{c,poi}$ is the shortest travel time from centroid $c$ to point of interest $poi$ and $T_\text{max}$ is a predefined threshold for travel time (e.g. 30 mins).
Intuitively, $acc(c)$ measures the number of PoIs that can be reached by individuals departing from centroid~$c$, within time~$T_\text{max}$.
Such PoIs are weighted by the time to reach them, so that the closer a PoI, the more it contributes to accessibility. Our definition is a combination of two classic definitions of accessibility, namely the \emph{isochrone} and \emph{gravity-based}~\cite{miller2020measuring}.
The purely isochrone definition of accessibility has the issue of counting all PoIs the same, despite the different in travel time to reach time. On the other hand, the purely gravity based definition of accessibility, factors in all PoIs, even those that would require a prohibitive travel time. By combining the two aspects, we solve the aforementioned limitations.

We define the global accessibility of graph $\pazocal{G}$ as 
\begin{equation}
    acc(\pazocal{G}) = \sum_{c \in \pazocal{C}}{acc(c)}. 
\label{eq:acc_graph}
\end{equation}
Classic efficiency-based optimization of PT would aim to maximize~$acc(\pazocal{G})$. Instead, our aim is to reduce the inequality in the geographical distribution of accessibility, and thus we choose to consider the accessibility of the centroids that suffer from the worst accessibility, as we aim to concentrate improvement in such zones.

One could be tempted to apply max-min optimization, trying to maximize the lowest accessibility in the territory. However, when we tried that, we obtained poor results. Indeed, we were ending up improving areas that were remote and often uninhabited. Often, the improvement was enjoyed by too few locations. We therefore propose to maximize some bottom quantile of the accessibility distribution. To the best of our knowledge, this simple yet effective idea has not been explored so far. We define the following accessibility metric, related to the $q$th quantile:

\begin{equation}
    acc^q(\pazocal{G}) = \sum_{c \in \pazocal{C}^q}{acc(c)}
\label{eq:acc-q}
\end{equation}

where $\pazocal{C}^q$ defines the set containing the $q$\% of centroids with the least accessibility.
Note that $acc(\pazocal{G})=acc^{100}(\pazocal{G})$.

\subsection{Problem definition}
\label{sec:optimization-problem}
Let us consider a PT graph $\pazocal{G}$ and a set~$\pazocal{B}$ of $n_b$ \emph{candidate bus stops}. Set~$\pazocal{B}$ is contained in set~$\pazocal{V}$ of nodes of $\pazocal{G}$. Set $\overline{\pazocal{B}}=\pazocal{V}\setminus\pazocal{C}\setminus\pazocal{B}$ is the set of \emph{non candidate stops}, i.e., the ones that will not be used to create the new lines.\\ In broad terms, we consider the problem of the PT operator
to design $k$ bus lines $\{l_1,\dots,l_k\}$ (where number~$k$ is fixed a-priori), passing by these $n_b$ bus stops, in order to reduce inequality of accessibility.
Any stop in~$\pazocal{B}$ may be already part of pre-existing lines or not. The problem at hand may emerge in case a PT operator wishes to build additional bus lines, passing by stops~$\pazocal{B}$. Another case is when a PT operator wishes to redesign current bus lines, while reusing current bus stops.

In quantitative terms, we wish to find graph $\pazocal{G}^*$, which is as $\pazocal{G}$ but also contains additional lines $l_1,\dots,l_k$, such that $acc^q(\pazocal{G}^*)$ is maximized.

\section{Resolution method based on Graph Reinforcement Learning}

We decompose our problem (\S\ref{sec:optimization-problem}) as a bilevel optimization: in the upper level, we partition the candidate bus stops in~$k$ subsets. In the lower level, each subset will be transformed in a line, deciding the order of the stops.

\begin{figure*}[h]
  \centering
  \includegraphics[width=0.7\linewidth]{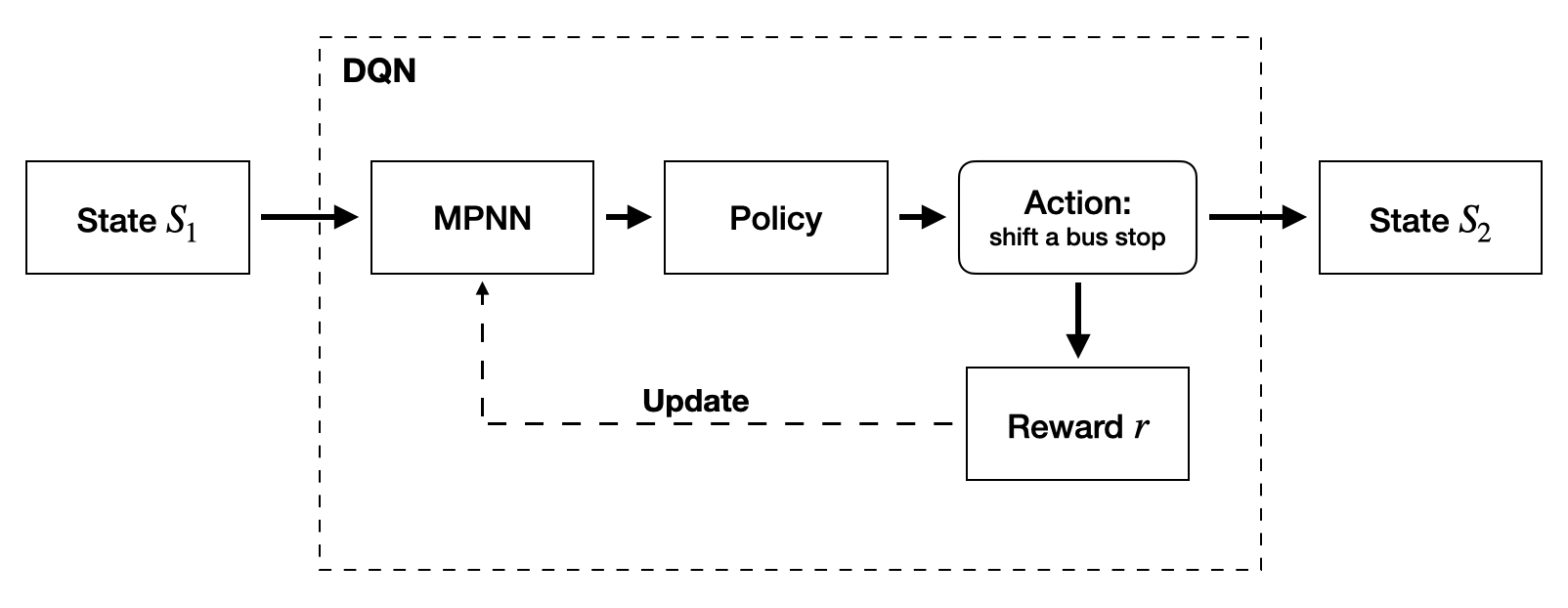}
  \caption{Framework.} \label{Framework}
\end{figure*}

\subsection{Markov Decision Process formulation}
\label{sec:RL_formulation}
\label{sec:markov-decision-process-formulation}
Let us denote by~$\pazocal{G}$ the initial graph, i.e., the one when no new bus lines have yet been added. We model the upper level problem as the following Markov Decision Problem (MDP):
\begin{itemize}
    \item \textbf{States}: A state is a partition~$S=(l_1, \dots, l_k)$ of candidate stops, where~$l_i=\{b_{i_1},\dots,b_{i_{n^i} }\}$
    is the set of bus stops assigned to line $l_i$.\footnote{For simplicity of notation, we use the same symbol $l_i$ to denote a line and also the set of stops assigned to it.}
    Each bus stop is assigned to a single line. 
    Given any state $S$, we build the line corresponding set~$l_i$ of stops, $i=1,\dots,k$. To transform set~$l_i$ in a line we need to decide the order in which the stops in~$l_i$ will be visited. Such an order is calculated via a heuristic (Section~\ref{sec:sorting-algorithm}).
    Given sate~$S=(l_1, \dots, l_k)$ and having defined the lines corresponding to $l_i, i=1,\dots,k$, we add to graph~$\pazocal{G}$ the edges corresponding to those lines and obtain a new graph~$\pazocal{G}(S)$.
    \item \textbf{Actions}: At each step, our optimization agent shifts a bus stop $b_i$ from its current line $l_o$ to a target line $l_t$. The action is defined by a tuple $a=(b_i, l_t)$. The state changes from~$S$ to~$S'$: state~$S'$ is equal to $S$, except for line~$l_o$ which becomes~$l_o=l_o\setminus \{b_i\}$, and for $l_t$, which becomes~$l_t=l_t\bigcup \{b_i\}$. The action of changing a bus stop to its own line is not admitted. Observe that ours is a Deterministic MDP~\cite{Hazan2013}, i.e., arrival state~$S'$ can be deterministically calculated from departing state~$S$ and action~$a$.
    \item \textbf{Rewards}: The instantaneous reward collected when applying action~$a=(b_i, l_t)$ on state $S$, is $$r(S,a)=acc^q\left(\pazocal{G}(S')\right) - acc^q\left(\pazocal{G}(S)\right)$$, where parameter~$q$ must be chosen in advance.
    \item \textbf{Policy}: During training, our agent follows an $\epsilon$-greedy policy. At test time, actions are chosen greedily with respect to the Q-values but our agent keeps exploring with a random action everytime it finds a local optima.
\end{itemize}

The sizes of the state space and the action space are~$k\cdot n_b$, considering a matrix with $k$ lines and $n_b$ bus stops can represent any state and action.

\subsection{High-level view of the optimization approach}


Due to the high size of the space and state spaces, enumerating all the states and actions and learn a $Q$-function that takes directly those states and actions as input is hopeless. Therefore, as common in graph-related optimization tasks, we resort to a Message Passing Neural Network~(MPNN)~\cite{LeiChen2023,gilmer2017neural,hameed2023graph,Kutyniok2022}. Via a MPNN, we embed each node in a low dimension Euclidean space. Such representation captures the ``role'' of that node within the graph, based on the direct or indirect connections with the other nodes. The process of node embedding is thus able to capture the structure of a graph, so that the RL agent can take decisions that take such structure into account.

Figure~\ref{Framework} depicts our framework. A MPNN takes as input $\pazocal{G}(S)$ PT graph of current state~$S$, then it outputs the Q value for each action.
Next, the Greedy Policy performs an action according to the Q values. At last, the reward, which is the change of accessibility metric, helps to update parameters of MPNN. The following section will introduce MPNN in more detail. Every time we shift a bus stop from a bus line to another line, a reasonable method of deleting and inserting bus stops in a line is presented in Section~\ref{sort}.

\subsection{Message Passing Neural Network}
\label{MPNeural_Network}

Let us associate to each candidate stop~$b$ a feature vector~$x_b$.
Let us denote with $X$ the matrix of the feature vectors of all stops and with $M_\text{adj}$ the adjacency matrix, where element $(i,j)$ is $1$ if there is a line in which bus stop $b_i$ comes right before $b_j$, and $0$ otherwise.


A MPNN calculates a vector~$\mu_b$, called \emph{embedding}, for each candidate bus stop~$b$.
Embedding~$\mu_b$ is a function of feature vector~$x_b$ , of feature
matrix~$X$ and of adjacency matrix $M_\text{adj}$.

In the following, all vectors like $\Theta_j$ denote parameters that learned during training. The embedding of bus stop $b$ is initialized as 


\begin{equation}
    \mu_b^0=f(x_b, \Theta_1) \in \mathbb{R}^{n}
\end{equation}
The vector of edge embeddings, one per each edge, is initialized as
\begin{equation}
    w^0 = g(M_\text{adj}, X, \Theta_2)
     \in \mathbb{R}^{m}
\end{equation}
Let us denote with $\pazocal{N}(b)$ the neighbor bus stops of bus stop $b$.
Note that since the bus lines change from a state~$S$ to another, $\pazocal{N}(b)$ may also changes for bus stop~$b$. 


The information is then shared to each of the nodes' neighbors through $T$ rounds of messages. The message passed to candidate stop~$b$ at round~$t+1$ is:

\begin{equation}
    m_b^{t+1} = M(\mu_b^t, \{\mu_u^t\}_{u\in \pazocal{N}(b)}, \{w^t_{ub}\}_{u\in \pazocal{N}(b)\}}, \Theta^{t}_3)  \in \mathbb{R}^{n'},
    \label{eq:M}
\end{equation}
where $w^t_{ub}$ is the embedding of edge between bus stop~$u$ and bus stop~$b$ after~$t$ times iterations. 

Embedding~$u_b^{t+1}$ of candidate stop $b$ by message:
\begin{equation}
    \mu_b^{t+1} = U(\mu_b^t, m_b^{t+1}, \Theta^{t}_4)
     \in \mathbb{R}^{n},
     \label{eq:U}
\end{equation}
$M$ and $U$ are respectively message and update functions at round $t$.
%
After~$T$
rounds of message passing, a prediction is produced by some readout function, $R$. In our case, the prediction is the set of Q-values corresponding to the actions of the changing bus stops and their target line:
\begin{align}
    \{Q(S,a)\}_{a \in \pazocal{A}} = R(\{\mu_b^T\}_{b\in \pazocal{B}}, \Theta_5)
    \label{eq:Q}
\end{align}
where $\pazocal{A}$ is the set of actions and $\pazocal{B}$ is the set of bus stops.
Note that $\mu_b^T$ is calculated via formula~\ref{eq:M} and formula~\ref{eq:U}, thus varys for different state~$S$.
Message function $M$, Update function $U$ and Readout function $R$ , as well as the embedding functions $f$ and $g$, are all neural network layers with learnable weights $\{\Theta_1,\Theta_2,\{\Theta^{t}_3\}_t,\{\Theta^{t}_4\}_t,\Theta_5\}$. Every time our agent takes an action and gets a reward~$r$, and One-Step Q-learning loss is:
\begin{align}
    Loss(S,a) = (\gamma \cdot max_{a'} Q(S',a') + r - Q(S',a) )^2,
    \label{eq:loss}
\end{align}
where $\gamma$ is a Discount factor. We update learnable weights via SGD.

\subsection{Sorting algorithm}
\label{sort}
\label{sec:sorting-algorithm}
Recall that state~$S$ is a partition~$(l_1,\dots,l_k)$ of set~$\pazocal{B}$ of candidates nodes. State~$S$ just establishes to which line each candidate node belongs. However, to transform any set~$l_i$ into a line, a certain ordering of its stops must be established.

In this section, we will introduce the method of determining the order of bus stops.
This optimal method should also maximize our accessibility objective function. However,since it is a Traveler Salesman Problem, defining such a function seems utterly complex or with a very high computational cost. We decided to use the shortest path algorithm as a proxy for the maximum accessibility path algorithm. 
We understand that this approach is suboptimal as we are optimizing distances and accessibility is a measure completed over the whole graph. Nevertheless, using this algorithm optimizes the headway of our bus lines and thus renders a good enough accessibility with low computational costs. 

\subsection{Reinforcement Learning Equality algorithm}

Given the number of bus line $k$ and PT network $\pazocal{G}$, we propose a Reinforcement Learning Equality algorithm (Algorithm~\ref{fig:algo_2}) to plan $k$ bus lines efficiently, in order to optimize accessibility objective function $acc^q(\cdot)$ ($ q = 20, 50, 100.$).
Our base idea is to get a better state-action function $Q$ by updating the MPNN network. Note that only when our objective function value is better than before, we will update our MPNN network. If $acc^q(\pazocal{G})$ has not changed in the past certain iterations (e.g. 5 iterations), the episode will be ended. Algorithm~\ref{fig:algo_2} terminates when a certain time threshold (e.g. 1h) is exceeded. In fact, the initial graph $\pazocal{G}$ of our algorithm can be different for each episode. Therefore, it has the ability to learn among different PT networks. 



\begin{algorithm}
    \caption{Online Reinforcement Learning Equality algorithm}
	\begin{algorithmic}[1]
        \STATE \textbf{Input}: Number of lines $k$, initial graphs $\pazocal{G}$, quantile~$q$.
        \REPEAT
		\STATE \textbf{Initialization}: \STATE  Randomly partition the bus stops between the lines to obtain initial state~$S$. 
            \STATE Sort the lines (\S\ref{sec:sorting-algorithm}), and update $\pazocal{G}$.
            \STATE  Set $acc^q_{best}=acc^q(\pazocal{G})$. 
    		\REPEAT
            \STATE Predict $Q=MPNN(\pazocal{G}, S)$ \eqref{eq:Q}
            \STATE Update the state with action $a=\argmax_{a} Q(S,a)$
            \STATE Sort each line $l_i$ (\S\ref{sec:sorting-algorithm})
            \STATE Calculate headway $t_{l_i}$ via~\eqref{eq:tl} for each $l_i$
            \STATE Update graph $\pazocal{G}$ with the new bus lines and headway
            \STATE Compute $acc^q(\pazocal{G})$ (see~\eqref{eq:acc-q})
            \IF{$acc^q_{best}>acc^q(\pazocal{G})$} 
                \STATE Set $acc^q_{best}=acc^q(\pazocal{G})$
                \STATE Compute loss via~\eqref{eq:loss}
                \STATE Update learnable weights $\{\Theta_1,\Theta_2,\{\Theta^{t}_3\}_t,\{\Theta^{t}_4\}_t,\Theta_5\}$ of MPNN by Gradient Descent
            \ENDIF
    		\UNTIL $acc^q(\pazocal{G})$ has not changed in the past $5$ iterations
        \UNTIL Running time threshold is exceeded
		\STATE \textbf{Return} The best accessibility $acc^q_{best}$
	\end{algorithmic}  
 \label{fig:algo_2}
\end{algorithm}

\section{Evaluation}
\label{Evaluation_}

\begin{figure}[h]
  \centering
  \includegraphics[width=0.6\linewidth]{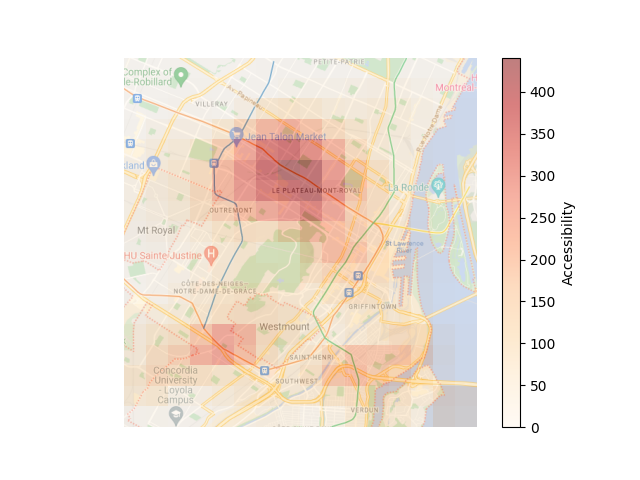}
  \caption{Accessibility of Montreal Metro network.} \label{fig:teaser}
\end{figure}

\todoi{aa: In all figures where ``Accessibility'' is present, either in the colorplot or any other plot, the units of measurement should appear}

\subsection{Considered scenario}

To evaluate our agent, we consider a simplified version of Montreal. We assume the PT operators wishes to keep the metro (subway) network as it is, but wishes to build bus lines in order to reduce the inequality of the geographical distribution of accessibility. Therefore, set~$\pazocal{B}$ of candidate stops consists of all bus stops, while set~$\overline{\pazocal{B}}$ of non canidate stops are the metro stops. Figure~\ref{fig:teaser} shows the accessibility distribution of accessibility resulting from initial PT graph~$\pazocal{G}$ of Montreal, which we assume  consists of only the current lines. These assumptions may correspond to the case in which the PT operator wishes to completely redesign bus lines (so that we can remove all bus lines from our initial graph~$\pazocal{G}$). The metro network is composed of 4 lines of different sizes. It covers mostly the center of Montreal, which delimits our environment boundaries. We do not consider the real locations of bus stops in Montreal, as this would be outside of the scope of this paper. To establish preliminary results, we limit at first the number of bus stops to an arbitrary low number of 72. We generate the bus stops as follows. We tessellate the territory with a regular grid. Note that this tessellation is not necessarily the same as the one described in Section~\ref{territoryPT_}. Within each cell of such a grid a bus stop is created with a random location within the cell. By generating bus stop locations in this way, we ensure uniformity of the bus stops at a larger scale. Consequently, all areas of the territory have access to the public transport network to all regions. We extract points of interest from Open Street Map \cite{OpenStreetMap} in combination with the Overpass API and assign to the centroids the point of interest number in their area. For the points of interest, we select some of the main amenities in a city (schools, hospitals, police stations, libraries, cinemas, banks, restaurants and bars). Some of these amenities present a distribution which is relatively uniform over the map (e.g. schools) while some others have higher densities in popular streets (e.g. restaurants). Scenario parameters are in Table~\ref{tab:param}.


\begin{figure*}[h]
  \centering
  \includegraphics[width=0.8\linewidth]{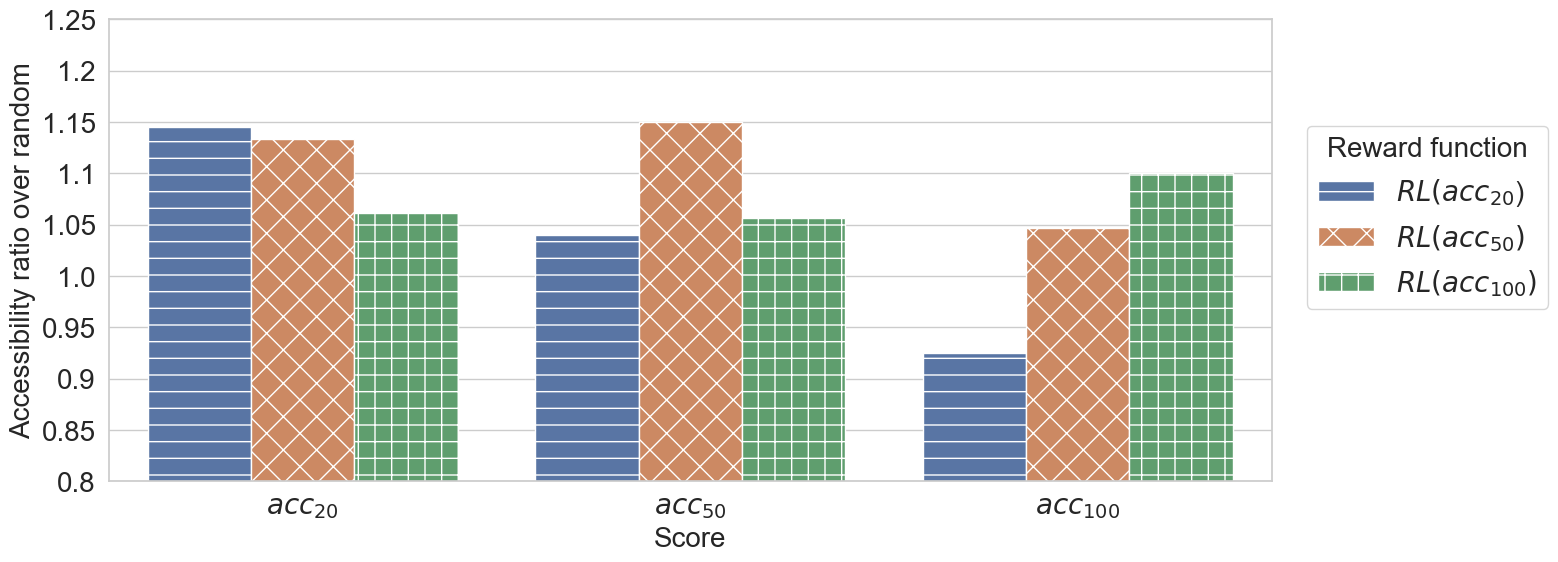}
  \caption{Different metric results via our Reinforcement Learning Equality algorithm against the random search algorithm.} \label{Basebench}
\end{figure*}

\begin{table}[]
\centering
  \caption{Scenario parameters}
  \label{tab:param}
  \begin{tabular}{cc}
    \toprule
    Parameter &Value\\
    \midrule    
    Number of bus stops $n_{bs}$& 72\\
    Number of lines $k$& 3\\
    Maximum accessibility time $T_\text{max}$& 30 minutes\\
    Walking speed $s_w$ \cite{ali2018case}& 4.5 km/h\\
    Bus speed $s_b$ \cite{ishaq2020designing}& 28km/h\\
    Fleet size per line& 10\\
    Distance between adjacent centroids& 1km\\
    Discount factor $\gamma$ & 0.95\\
    \bottomrule
  \end{tabular}
\end{table}


\subsection{Baselines}
We compare the performance of our algorithm to two baselines. To compare the algorithms, we define a maximum running time of 1-hour and check which is the best bus lines deployment found.

\subsubsection{Random search algorithm}
Random states of the form~$S=(l_1,\dots,l_k)$ are generated, each corresponding to a random partition of set~$\pazocal{B}$ of candidate stops. At every generated state~$S=(l_1,\dots,l_k)$, each set~$l_i$ is sorted as in Section~\ref{sec:sorting-algorithm}
to generate the corresponding lines. Then the corresponding graph~$\pazocal{G}^\text{rnd}$ is constructed and accessibility~$acc^q(\pazocal{G}^\text{rnd})$ is computed. This process is repeated until the running time threshold is exceeded. The value of~$acc^q(\pazocal{G}^\text{rnd})$ is then returned.
This random search algorithm has the advantage over our algorithm to visit very diverse states and no time for computing node embeddings, rewards, etc. The number of states visited by this algorithm is much larger than the one visited by our approach, within the same running time threshold. 


\subsubsection{Genetic algorithm}
The second baseline is a genetic algorithm (Algorithm~\ref{fig:algo_GA}). It is a popular metaheuristic to solve combinatorial problems, and PT design problems in particular \cite{farahani2013review}.
We adopt~$acc^q(\pazocal{G})$ as the fitness function. Next, the design of the evolutionary functions must be done carefully so that the new generations inherit good genes from their parents. Therefore, the most important function that must be modified for that case is the crossover function. 
More popular approaches, like Order Crossover 1 (OX1) preserve relative order of the states and thus their structure, are used in similar problems like TSP \cite{kora2017crossover} and VRP \cite{prins2004simple}. Children can benefit from advantageous orderings of the parent nodes, as from our experience, close nodes (sorted by the $SORT$ function) are often good to increase accessibility.

\begin{algorithm}
    \caption{Genetic algorithm}
	\begin{algorithmic}[1]
        \STATE \textbf{Input}: Size $n_{pop}$ of the population, number of lines $n_{lines}$, number of parents $n_{par}$ per generation, timeout $t_{limit}$ of the benchmark, probability $P_{mut}$ of an attribute to mutate.
		\STATE \textbf{Initialization}: Initialize $pop_0 = (I_1, \dots, I_{n_{pop}})$ at generation $0$ as individuals with random partitions of the bus stops between the lines. Sort the lines of each individual. Compute $acc^{best}_0$
        \REPEAT
            \STATE Select $n_{par}$ best $parents= Tournament(pop, acc_{20})$
            \REPEAT
                \STATE $p_1, p_2 = Sample(parents, 2)$
                \STATE $c = OX(p_1, p_2)$
                \STATE $c = Mutate(c, P_{mut})$
                \STATE Sort the lines of each child $c=SORT(c)$
                \STATE $pop_i.Append(c)$
            \UNTIL $pop_i$ is filled with $n_{pop}$ children.
            \STATE Compute $acc^{best}_{i} = \max(acc^q(pop))$ 
        \UNTIL The end of the benchmark
		\STATE \textbf{Return} The best accessibility over all generations $acc^{best}$
	\end{algorithmic}  
 \label{fig:algo_GA}
\end{algorithm}

\subsection{Results}


For the first comparison, we train our Reinforcement Learning Equality agent and test it on the same bus stop distribution. The random search algorithm \ref{fig:algo_2} baseline is defined with different accessibility metric. Figure~\ref{Basebench} shows that our approach outperforms the random baseline by 10\%-15\% when trained and tested on the same accessibility metric (e.g. the agent optimizes $acc_{20}$ with its reward function and is tested on the same metric). Our algorithm also improves the other different accessibility metrics than the one it has been trained by an average ratio of 5\% over the baseline. Only the agent trained on $acc_{20}$ performs worse than the random baseline. Because when we optimize PT with equality ($acc_{20}$) as the optimization goal, we may lose some efficiency ($acc_{100}$). Planning bus lines need to make a trade-off between equality and efficiency.

\begin{figure}[h]
  \centering
  \includegraphics[width=0.6\linewidth]{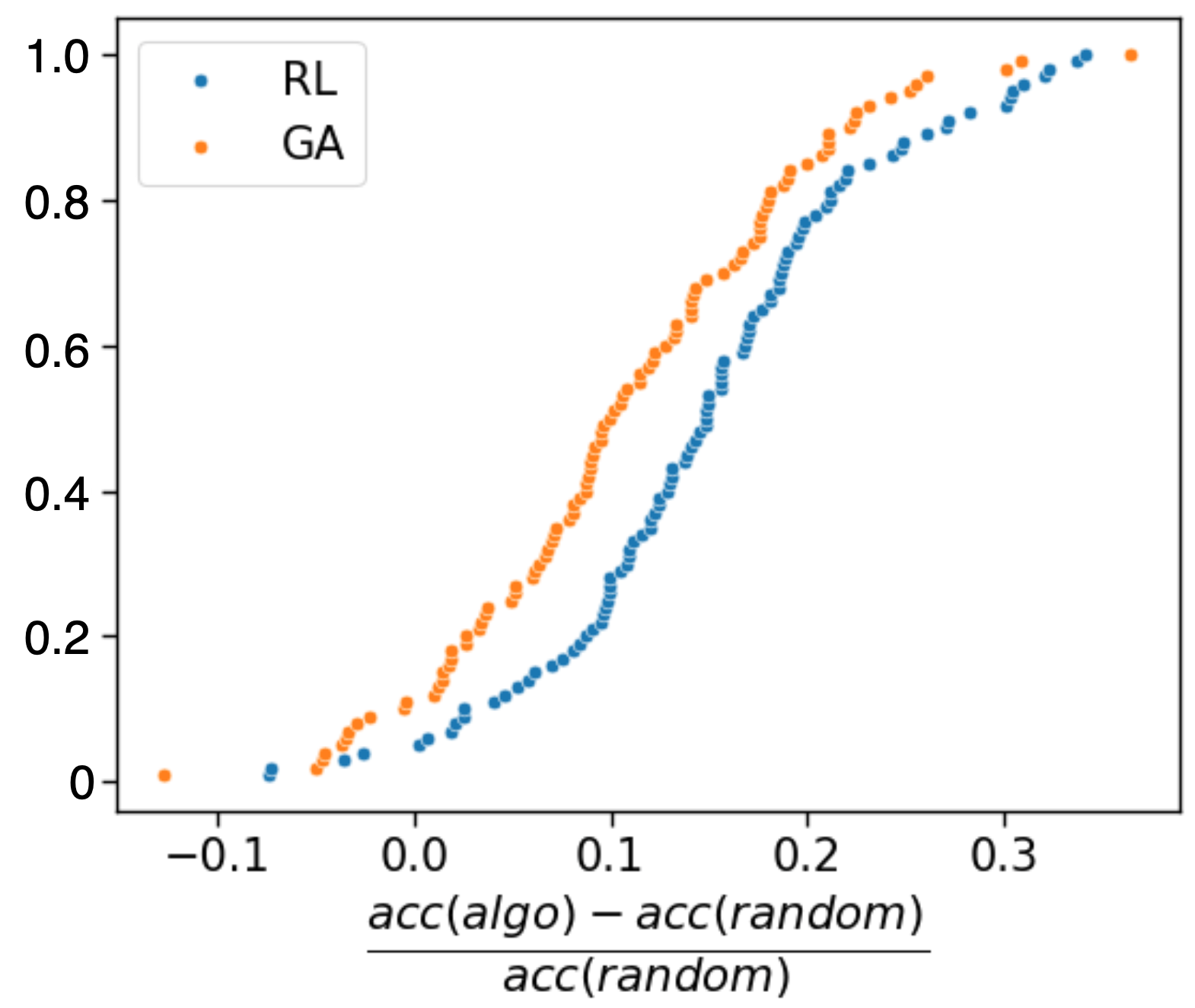}
  \caption{ CDF of $acc_{20}$ improvements of different algorithms comparing with random baseline.} \label{fig:Ttest}
\end{figure}

\begin{figure}
  \centering
  \includegraphics[width=0.6\linewidth]{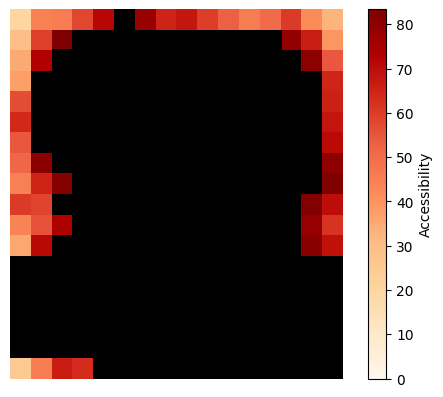}
  \caption{Heatmap of improvement of accessibility via Reinforcement Learning Equality algorithm against the random baseline on underserved areas (The darker red areas indicate that our approach improves more obviously than the random baseline in these areas. Black areas with the best 80\% accessibility are not considered).} \label{fig:Colormap}
\end{figure}

Our agent is also able to apply to different bus stop distributions. We assume that each bus stop remains in the same grid cell, but its position is changed for different distributions. Now we select $acc^{20}$ as our accessibility metric for training and testing on a set of 100 PT graphs with different bus stop distributions. 
In fig~\ref{fig:Ttest}, we plot the CDF of $acc_{20}$ improvements value $R = \frac{acc(algo)-acc(random)}{acc(random)}$ where $acc(random)$ is the optimized $acc^{20}$ value by the random search and $acc(algo)$ is the optimized $acc^{20}$ value by Algorithm~\ref{fig:algo_2} or GA. We observe in both cases that $R$ follow a normal distribution. Using this data, we conduct a t-test with the null hypothesis ($H_0$) that $mean(R)=0$. We run the t-test and get a T-statistic value of $17$. From the p-value $p=3*10^{-31}$, we can reject $H_0$ with confidence more than $99.9\%$. Thus, we confirm that our approach work better than random baseline on graphs with different bus stop distributions.
Similarly, the Genetic Algorithm performs better than the random baseline. Using a similar procedure to compare our Reinforcement Learning Equality algorithm and the Genetic algorithm, we show that our approach performs better on different test graphs with confidence $>99.9\%$ (the T-statistic value is $8$). 


We next test the performance of our algorithm on the actual geography of Montreal. Considering $acc^{20}$ as the objective function, so our algorithm focuses more on optimizing those areas with the 20\% worst accessibility. We also calculate the value of improvements compared to Random Baseline ($acc^{20}_\text{RL} - acc^{20}_\text{Random}$) for each centroid with the lowest accessibility in the initial graph. Figure~\ref{fig:Colormap} shows that our approach preform better than random baseline in these areas, the improvement in most of these areas exceeded 40\%. 

\section{Conclusion}

In this paper, we proposed an approach that combined Message Passing Neural Networks (MPNN) and Reinforcement Learning (RL) to optime bus line design. The objective is to reduce the inequality in the distribution of accessibility provided via Public Transport (PT). Our results showed that MPNN and RL are more effective than commonly used metaheuristics, as they can capture the PT graph structure and learn the dependencies between lines and the dsitribution of Points of Interest, where metaheursitcs restrict themselves to a random exploration of the solution space.

In future work, we will test how our method generalizes to different metro networks, realistically modeled via open data, and how it scales when the problem involves much more bus stops than considered here.
Moreover, we expect even better results when allowing the same bus stop to be part of multiple bus lines (which is not the case in this work).

Tha main takeaway of this paper is that combining MPNN and RL is promising to solve PT network design, at a strategic planning phase.


\newpage


\bibliography{sample-base}


\end{document}